# Analyzing Learned Convnet Features with Dirichlet Process Gaussian Mixture Models

**David Malmgren-Hansen**[*]
Department of Applied Mathematics and Computer Science
Technical University of Denmark
Lyngby, Denmark
`dmal@dtu.dk`

**Allan Aasbjerg Nielsen**
Department of Applied Mathematics and Computer Science
Technical University of Denmark
Lyngby, Denmark
`alan@dtu.dk`

**Rasmus Engholm**
Terma A/S
Lystrup, Denmark
`rae@terma.com`

## Abstract

Convolutional Neural Networks (Convnets) have achieved good results in a range of computer vision tasks the recent years. Though given a lot of attention, visualizing the learned representations to interpret Convnets, still remains a challenging task. The high dimensionality of internal representations and the high abstractions of deep layers are the main challenges when visualizing Convnet functionality. We present in this paper a technique based on clustering internal Convnet representations with a Dirichlet Process Gaussian Mixture Model, for visualization of learned representations in Convnets. Our method copes with the high dimensionality of a Convnet by clustering representations across all nodes of each layer. We will discuss how this application is useful when considering transfer learning, i.e. transferring a model trained on one dataset to solve a task on a different one.

## 1 Introduction

Convolutional Neural Networks (Convnets) have had a great impact on a range of computer vision problems such as image classification, object detection, image captioning etc. The ability to comprise image context by representing it with hierarchically ordered feature extractions makes Convnets suitable for scaling to large complex problems with many categories of data. The reasoning for this is that lower level features will be simple building blocks of describing many types of data, while higher order features are more specific to a certain context and makes it possible to separate images by predefined classes.

There are rarely any built-in constraints on the structure or location of internal representations in Neural Networks. The idea is to train a Convnet to be a "good" hierarchically ordered feature extractor from large quantities of data. Whether this actually happens and which internal nodes or layers become certain types of feature extractors are very important in order to understand the generalizability of a given model. It can also give insight into how features learned on one dataset can be transferred to other problems, known as Transfer Learning, a field that has been heavily explored within deep learning recently, e.g. in [15], [8], [17], [11].

Showing feature maps or node activations of a Convnet as images, such as in [7] is one simple way of gaining insight, but it leaves the user to interpret a large number of abstract feature maps.

---

[*]http://www2.compute.dtu.dk/~dmal/

Common Convnet architectures can easily have several hundreds of feature maps in just one layer of the network. In [6] it was shown that the first layer parameters resembled Gabor wavelet filter kernels. This was interesting for the fact that a the large amount of image data and the complex problem with a 1000 classes had forced the filters to become, what has long been known to be good image texture descriptors, [4], [10]. Showing filter parameters in RGB space stops being a straight forward procedure after the first layer of a Convnet though. In deeper layers each node has a number of filters corresponding to the number of channels (feature maps) of the previous layer and are therefore no longer a color specific representation like in the first layer, illustration in Figure 1. This has led to other recently proposed techniques such as in [17], [16], [9]. In [17] a method is proposed for back propagating an internal representation from one node of a Convnet back to input RGB-space. The technique shows easy interpretable visualizations of a node, but still leaves the task of picking which nodes to visualize to the user. If one has a labelled dataset, a way to pick interesting nodes could be to find the ones that reacts strongly to a group in the dataset, e.g. cats. Without labelled images though, it is hard to find nodes that are relevant.

In this paper we propose an alternative approach based on Dirichlet Process Gaussian Mixture Model (DPGMM) for clustering feature representations. The DPGMM follows the method in [3], and has the great advantage that it can cluster our feature representations without knowing the number of clusters as oppose to non-inference clustering like GMM. Our approach is to cluster representations for each layer individually, and do it either in a pixel wise manner for layers with high spatial dimensionality, i.e. early layer feature maps, or do the clustering over image examples for deep representations. Since DPGMM is an unsupervised clustering algorithm we do not need labels on our images to find relevant parts of the network to visualize.

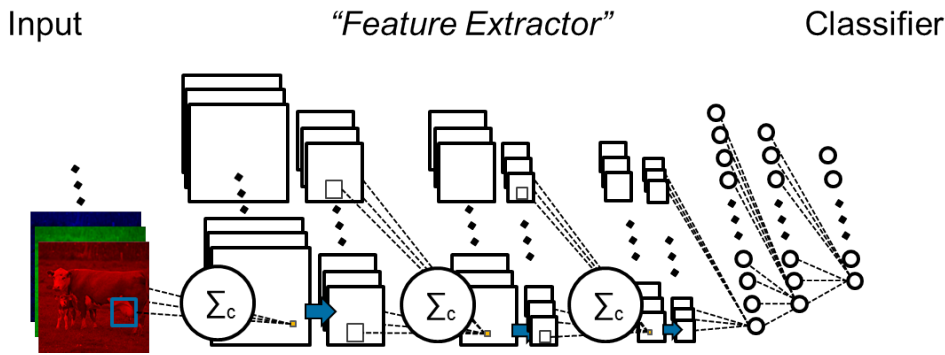

Figure 1: Example illustration of a Convnet (note the illustration above follows a different architecture than the one used in our experiments). Summation signs over c is sum over c input channels, squares denote feature maps from convolutional layers, blue arrows is the chosen subsampling scheme e.g. max-pooling and small circles are neurons in fully connected layers. Connections are only shown for the first node in each layer, but all nodes in all layers are connected to all inputs from previous layers.

## 2 Method

In Deep Learning it has become a standard to store input data and model parameters in N-dimensional arrays referred to as tensors, [2], [1]. This is especially convenient with Convnets when working on image data as your input will have four dimensions, i.e. number of images, number of channels (e.g. 3 for RGB), number of rows and number of columns. A convolutional layer output will be, for each image, a number of feature maps equal to the number of nodes chosen for the layer which each has a new number of rows and columns. In our experiment we analyze the pretrained Convnet VGGNet from [12] (model D), which predicts probabilities for each of the 1000 classes in ImageNet given a fixed size 224x224 RGB input image. The convolutional layers of this model produce feature maps with number of rows and columns equal to the number of rows and columns in the input, due to their choice of the convolutional layer hyper parameters called "stride" and "border mode". As the first layer in the model has 64 features maps the output size will be $(n, 64, 224, 224)$, with $n$ being the number of images we process with the network. The model has 16 layers and the feature maps' size is reduced trough the network with a max-pooling subsampling function in between layers. The last three layers are one-dimensional, i.e. fully connected layers. To cluster internal representations we need to consider them as points of features. Since we are interested in learning what each layer represents we suggest to cluster across nodes of each layer. For the fully connected layer this means each vector representation of an image is a point. For the convolutional layers' output (feature maps) we need to consider each pixel in them across all feature maps in the layer as a point. As an example we can consider the second layer in the model, which outputs an array of size $(n, 64, 224, 224)$. We will convert this into a matrix of size $(n \cdot 224^2, 64)$ considering each row a point in feature space. We then cluster the points with the DPGMM implementation from the Scikit-Learn package for Python and reconstruct the image of labels given from the assigned mixture components. This allows us to visualize a feature image where each color represents a certain cluster of features that repeats in the given layer.

The values of the feature map's pixels and fully connected layer activations are ranging in an arbitrary interval given by the learned parameters of the model. We scaled the values to range between 0 and 10 in order to fit the random initialization scheme of Gaussian Mixture components. Further we choose the maximum number of components to be 50 and from experiments found the alpha value of the Dirichlet Process prior of 0.2 to be a good choice.

## 3 Experiments

In our experiments we analyzed the dataset Warwick-QU presented in [13], [14], which consist of Haematoxylin and Eosin (H&E) stained slides of gland samples. Our goal is to explore the extent of Transfer Learning, in the sense that our dataset is very different from the ImageNet dataset ([5]) on which our model was trained.

It is possible to stack outputs of internal layers from any number of images, but when analyzing early layers with large output size, the number of points to cluster becomes large. In early layers it is possible to cluster representations from a single image though, whereas in fully connected layers we have to cluster across several images. In Figure 2 an example is shown with feature map clustering from early layers to deep layers.

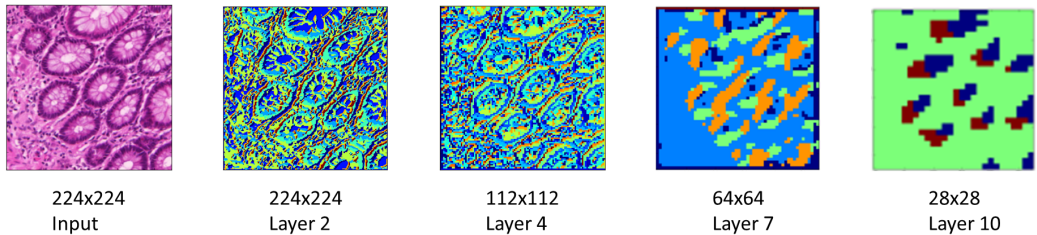

Figure 2: DPGMM clusters of feature maps from a single image for different layers in the network. Note that the feature map size decreases due to the Max-pooling subsampling operations.

It is seen in Figure 2 that the clusters of features represent larger areas from the input image as we go deeper into the network. An interesting aspect is also that the number of clusters gets lower, and in layer 10 the non-background clusters seems to represent a location of a cell in the input. We ran the algorithm on a number of cell images and plotted the center of clusters in its position in input space, shown on Figure 3. The point on the result on Figure 3 is not to present a good cell detector, but it is an interesting fact that it naturally arises from accumulated information in layer 10 of the pretrained Convnet. This shows that there is structured context representation all the way up to layer 10 in this model on this new dataset.

When analyzing the representations in even deeper layers there is little spatial information about them left in the representation. It is therefore necessary to look at clusters over a range of image examples. From layer 14 in our model, which is a fully connected layer, we have clustered all 165 images in the dataset. The algorithm clusters them as seen in Figure 4.

The Warwick-QU dataset is annotated with grades (benign, malign) and patient ID. Cluster "1" in Figure 4 is 74.4% benign image samples, whereas cluster "2" is 92.9% malign samples. Clusters "3", "4" and "5" single sample clusters, and sort of outliers in our clustering result, they are all benign samples. Cluster "6" has two benign samples from the same patient and cluster "7" has four malign samples from another patient. We conducted experiments with the concentration parameter alpha equal to 0.1 and 0.2. With alpha equal to 0.2 cluster "1" became 100% benign of size 27 samples, at the cost of cluster "2" only being 64% malign. That the clusters to some extent contains a meaningful grouping of the images, suggest that our algorithm can do clustering of unseen data, but also that features learned on natural images of objects can be useful on data recorded in a distinctly different setup. What our experiment shows is to what extent the learned features are useful from a given Convnet. The feature vector used to cluster the images in Figure 3 was extracted from layer 14 out of 16 layers. This suggest that a big part of the network generalizes well and is likely to increase performance when used for transfer learning purposes.

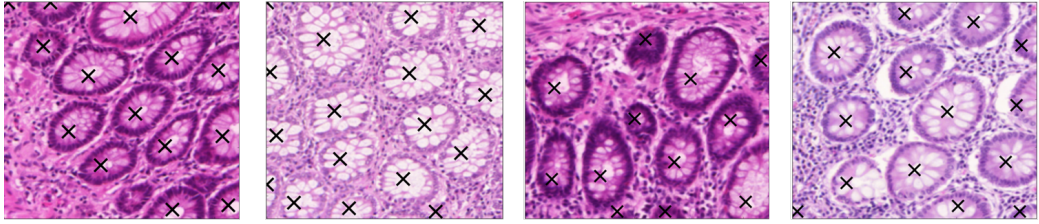

Figure 3: Non-background cluster centers from layer 10 plotted in input space for different cell images.

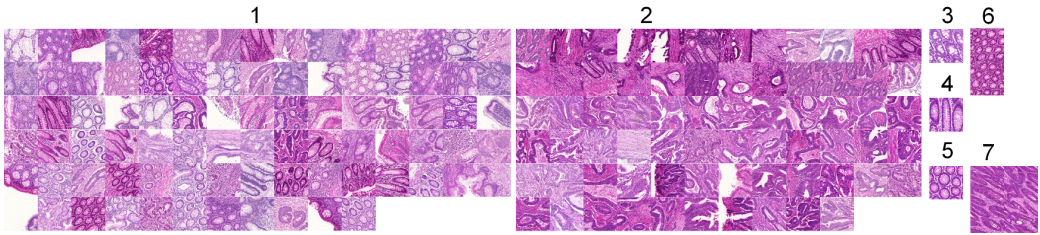

Figure 4: Clustering over all images of the vector representation from layer. Alpha value was 0.1 for the DPGMM clustering.

The original task of the Warwick-QU dataset was to segment the images. It is not in the scope of this paper to explore the process of using a Convnet model trained for classification on a segmentation task, but some approaches on how to do this is presented in [8]. Given our analysis it seems promising to use a pretrained model for the segmentation task even though the original dataset is significantly different from the Warwick-QU data.

## 4 Conclusion

In this paper we have proposed a technique for clustering and visualizing internal representations in a pretrained Convnet, based on Dirichlet Process Gaussian Mixture Models. Our clustering approach is unsupervised which makes is possible to interpret the results regardless whether labels are available for the data. Our method copes with the high dimensionality of representations in a single layer of a Convnet, by compressing it to a discrete label space that can be represented with a color for each label. This reveals the number of different feature clusters a model layer represents data with for a given image.

The proposed algorithm is well suited to explore and explain the cross domain generalizability, that has been experimentally shown in transfer learning research recently. In future work we aim to apply the method to a greater variety of datasets, in order to learn more about the internal representations of data in Convnets. Another interesting area would be to investigate the difference in clustering performance on datasets more similar to the original one and less similar datasets like the one presented in this article.